\def\year{2021}\relax
\documentclass[letterpaper]{article} 
\usepackage{aaai21}  
\usepackage{times}  
\usepackage{helvet} 
\usepackage{courier}  
\usepackage[hyphens]{url}  
\usepackage{graphicx} 
\usepackage{natbib}  
\usepackage{caption} 
\graphicspath{}
\usepackage{diagbox}
\usepackage{subfigure}
\usepackage{algorithmic}

\usepackage{amsmath}		
\usepackage{amsfonts}
\usepackage{graphicx}
\usepackage{booktabs}
\usepackage{multirow}
\usepackage{multicol}

\usepackage{autobreak}
\usepackage{makecell}

\usepackage[linesnumbered,ruled,vlined]{algorithm2e}
\usepackage{enumerate}
\usepackage[table,xcdraw]{xcolor}
\usepackage{adjustbox}
\usepackage[graphicx]{realboxes}
\usepackage{rotating, graphicx}
\usepackage{array}
\usepackage[switch]{lineno}  %
\usepackage[T1]{fontenc}

\title{Fever Basketball: A Complex, Flexible, and Asynchronized Sports Game Environment for Multi-agent Reinforcement Learning}

\title{Fever Basketball: A Complex, Flexible, and Asynchronized Sports Game Environment for Multi-agent Reinforcement Learning}
\author {

        Hangtian Jia,\textsuperscript{\rm 1}
        Yujing Hu, \textsuperscript{\rm 1}
        Yingfeng Chen, \textsuperscript{\rm 1}
        Chunxu Ren,
        \textsuperscript{\rm 1}
        Tangjie Lv,
        \textsuperscript{\rm 1}
        Changjie Fan,
        \textsuperscript{\rm 1}
        Chongjie Zhang
        \textsuperscript{\rm 2}
        \\
}
\affiliations {
    \textsuperscript{\rm 1} Netease Fuxi AI Lab,
    \textsuperscript{\rm 2} Tsinghua University \\
   \{jiahangtian, huyujing, chenyingfeng1, renchunxu, hzlvtangjie, fanchangjie\}@corp.netease.com,\\ chongjie@tsinghua.edu.cn,
}

\begin{document}


\makeatother

\maketitle

\begin{abstract}

The development of deep reinforcement learning (DRL) has benefited from the emergency of a variety type of game environments where new challenging problems are proposed and new algorithms can be tested safely and quickly, such as Board games, RTS, FPS, and MOBA games.
However, many existing environments lack complexity and flexibility, and assume the actions are synchronously executed in multi-agent settings, which become less valuable. We introduce the Fever Basketball game, a novel reinforcement learning environment where agents are trained to play basketball game. It is a complex and challenging environment that supports multiple characters, multiple positions, and both the single-agent and multi-agent player control modes. In addition, to better simulate real-world basketball games, the execution time of actions differs among players, which makes Fever Basketball a novel asynchronized environment. We evaluate commonly used multi-agent algorithms of both independent learners and joint-action learners in three game scenarios with varying difficulties, and heuristically propose two baseline methods to diminish the extra non-stationarity brought by asynchronism in \textit{Fever Basketball Benchmarks}. Besides, we propose an \textbf{i}ntegrated \textbf{c}urricula \textbf{t}raining (ICT) framework to better handle Fever Basketball problems, which includes several game-rule based cascading curricula learners and a coordination curricula switcher focusing on enhancing coordination within the team. The results show that the game remains challenging and can be used as a benchmark environment for studies like long-time horizon, sparse rewards, credit assignment, and non-stationarity, etc. in multi-agent settings.


\end{abstract}

\section{Introduction}
\begin{figure*}[t]
\centering
\subfigure[]{
\begin{minipage}[t]{0.33\linewidth}
\centering
\includegraphics[width=0.98\columnwidth]{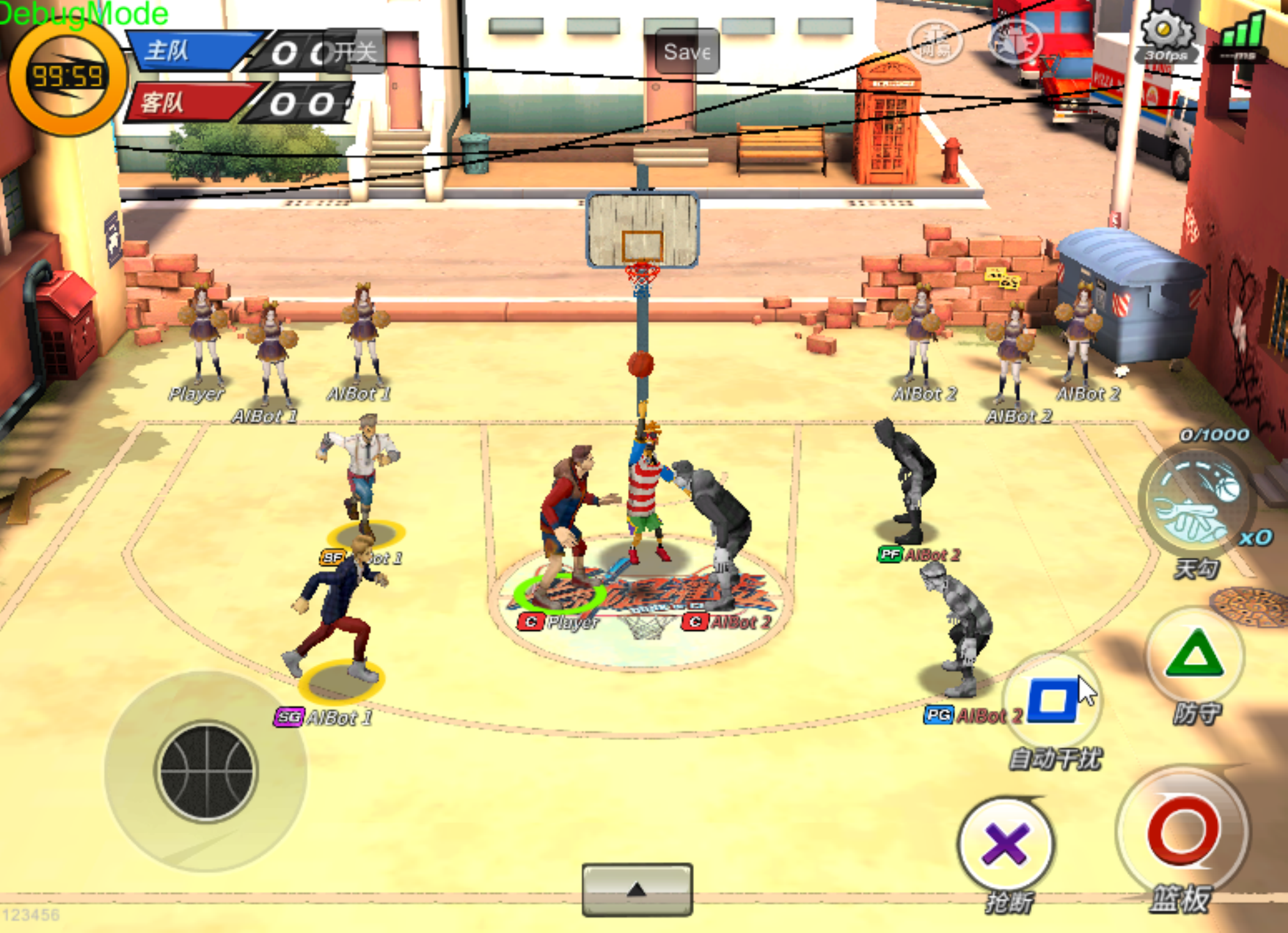}
\end{minipage}%
}%
\subfigure[]{
\begin{minipage}[t]{0.33\linewidth}
\centering
\includegraphics[width=0.98\columnwidth]{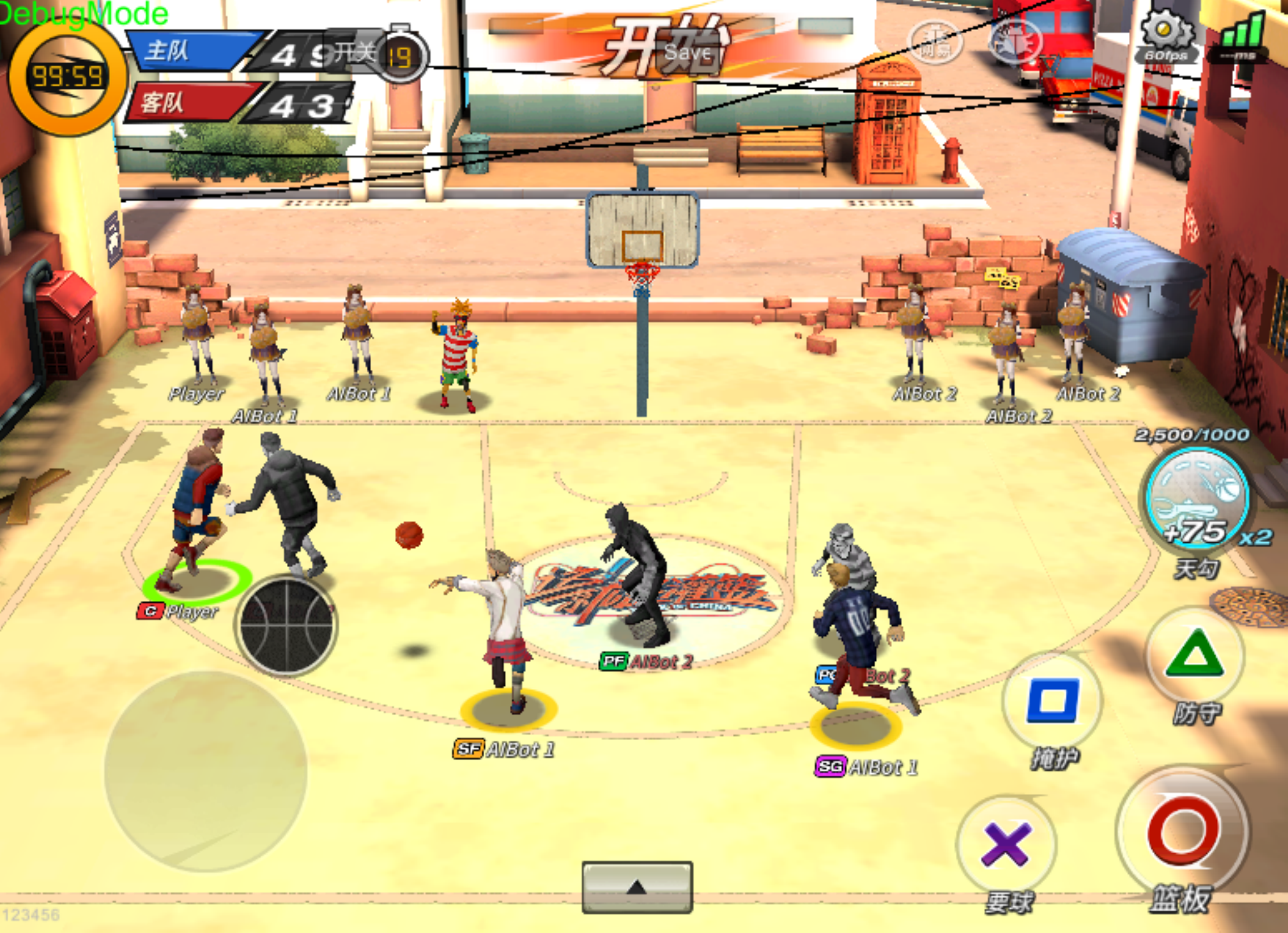}
\end{minipage}%
}%
\subfigure[]{
\begin{minipage}[t]{0.33\linewidth}
\centering
\includegraphics[width=0.98\columnwidth]{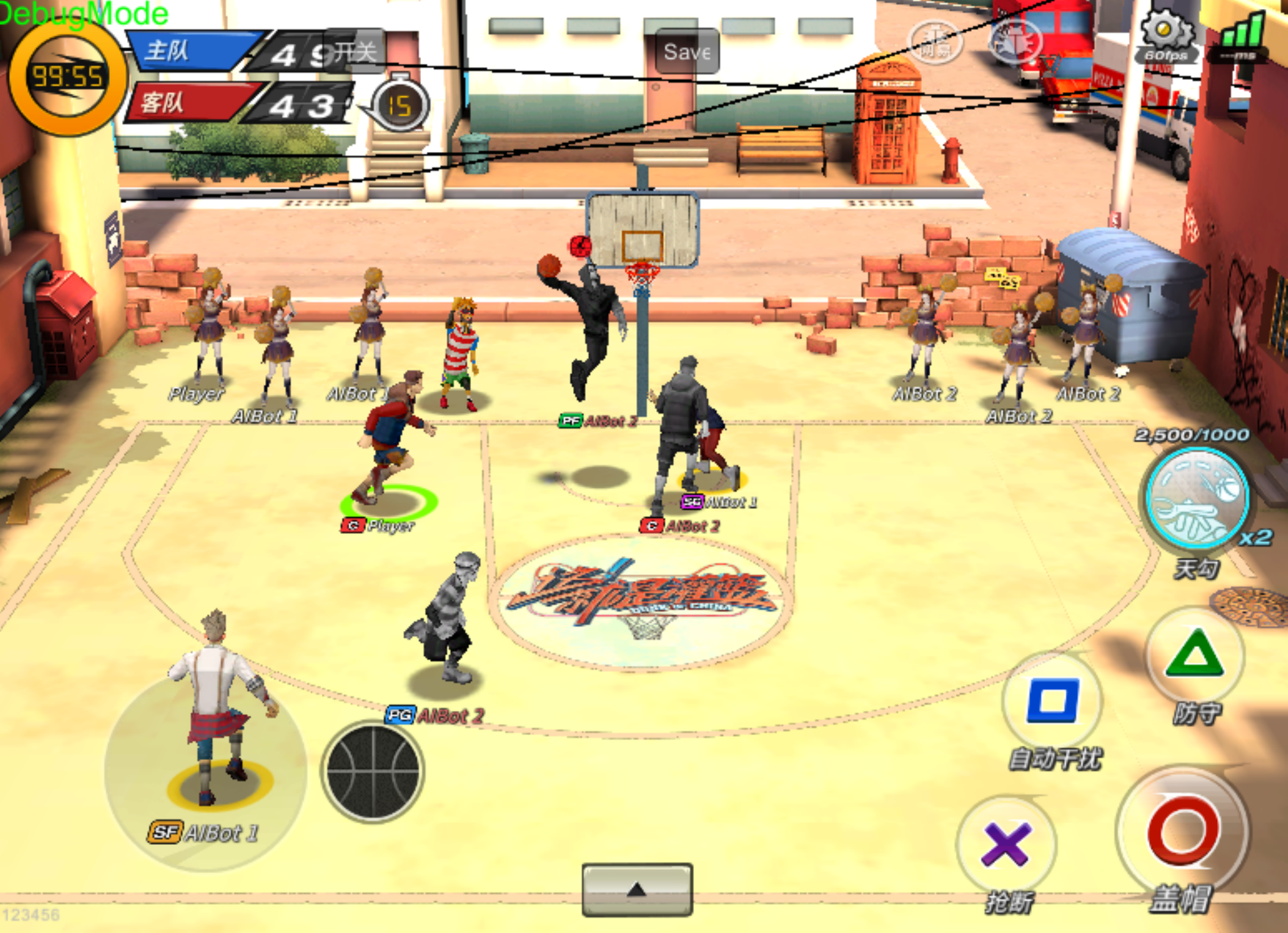}
\end{minipage}
}%

\centering
\vspace{-10pt}
\setlength{\belowcaptionskip}{-10pt}
\caption{Fever Basketball is a basketball simulator that supports major basketball rules and scenarios such as jump ball, offense, defense, passing ball, dunk, rebound, etc.}
\label{async_action}
\end{figure*}


Deep reinforcement learning (DRL) has achieved great success in many domains, including games \cite{mnih2013playing, silver2017mastering, lample2017playing}, recommendation systems \cite{munemasa2018deep}, robot control \cite{haarnoja2018soft} and autonomous driving \cite{pan2017virtual}. Among all these domains, games are one of the most active and popular settings. Because they are simulations of reality and have relatively lower cost of trial and error. Besides, games can be run in parallel to collect experience for training, which is another advantage of facilitating the success of DRL. 
There have been a variety kinds of game environments nowadays, for example, Atari games \cite{mnih2013playing, mnih2015human}, board games \cite{silver2016mastering, silver2017mastering}, card games \cite{heinrich2016deep}, first-person shooting (FPS) games \cite{lample2017playing}, multiplayer online battle arena (MOBA) games \cite{OpenAI_dota, jiang2018feedback}, real-time strategy (RTS) games \cite{vinyals2019alphastar, ijcai2019-631}. However, the lack of complexity and flexibility, and the assumption of synchronized actions in many existing environments that support multi-agent training remain potential barriers for better development of RL.

As a typical sports game (SPG), Fever Basketball simulates basic elements of basketball games (Figure $1$), which is challenging for modern RL algorithms. First of all, the long-time horizon and sparse rewards remain issues for most DRL methods. In basketball, it is normally not until scoring shall the agents get a reward (goal in or not), which may require a long sequence of consecutive events such as dribbling and passing the ball to teammates to break through the defense of opponents. Second, the whole basketball game is a combination of many challenging sub-tasks based on game rules, for example, the offense sub-task, the defense sub-task, and the sub-task of fighting for ball possession when the ball is free. Third, it is a multi-agent system that requires teammates to cooperate well to win the game. Fourth, players in reality have different reaction time, which makes the decision making asynchronized within the same team. Moreover, the different characters and positions classified according to players' capabilities or tactical strategies such as center (C), power forward (PF), small forward (SF), point guard (PG), and shoot guard (SG) add extra stochasticity to the game. All of these reasons described above make basketball a challenging SPG.

In this paper, we propose the Fever Basketball Environment, a novel open-source asynchronized reinforcement learning environment where agents can learn to play one of the world's most popular sports basketball. Building upon a commercial basketball game engine, our main contributions are as follows:

\begin{enumerate}[1)]
    \item We provide the Fever Basketball Environment, an advanced and challenging basketball simulator that supports all the major basketball rules.
    \item We provide the asynchronized Fever Basketball game clients to better simulate reality in multi-agent settings.
    \item We provide different training curricula (such as offense, defense, freeball, ballclear) for handling the whole basketball task.
    \item We provide various training scenarios (such as 1v1, 2v2, 3v3), multiple characters, and tasks of varying difficulties that can be used to compare different algorithms.
    \item We evaluate common algorithms for multi-agent scenarios and propose two heuristic methods for handling the asynchronism for joint-action learners.
    \item We propose an integrated curricula training (ICT) framework that reaches up to 70\% win-rate during a 300-day online evaluation with human players.
\end{enumerate}

\section{Motivation and Related Works}
The development of algorithms benefits from the emergence of new challenging problems, and game environments have been serving as the fundamental place where the reinforcement learning community tests its ideas nowadays. However, most of the existing environments have certain deficiencies which can be made up by Fever Basketball game:

\subsubsection{Low task complexity.}
As deep reinforcement learning algorithms become more sophisticated, existing environments with low task complexity and randomness become less challenging, and the benchmarks based on them become less informative \cite{juliani2018unity}. For example, the canonical \textit{CartPole} and \textit{MountainCar} tasks \cite{sutton2018reinforcement} are too simple to distinguish the performance of different algorithms. Meanwhile, most of the agents of \textit{Atari} games in the commonly used \textit{Arcade Learning Environment} \cite{bellemare2013arcade} have been trained to super-human level \cite{badia2020agent57}. The same applies to the \textit{DeepMind Lab} \cite{beattie2016deepmind} and \textit{Procgen} \cite{cobbe2019leveraging}, the former of which consists of several relatively simple first-person navigation maze environments and the latter is a suite of several game-like environments mainly designed to benchmark generalization in reinforcement learning. Besides, games in \textit{OpenAI Retro} \cite{nichol2018gotta} such as Sonic The Hedgehog can be easily solved by existing algorithms \cite{schulman2017proximal, hessel2018rainbow}.


\subsubsection{Fixed number of agents.}
Many existing environments only support the controlling of a fixed number of agents, and most of them are single-agent reinforcement learning (SARL) problems. For example, the \textit{Hard Eight} \cite{paine2019making} environment focuses on a single agent's training to solve hard exploration problems. The \textit{Obstacle Tower Environment} \cite{juliani2019obstacle} also only supports the training of a single agent to solve puzzles and make plans on multiple floors. The \textit{Atari} games and \textit{OpenAI Retro} games are also environments that support single-agent training.
However, most of the real-world scenarios involve more than one agents, such as basketball matches and autonomous driving cars \cite{shalev2016safe}, which can be naturally modeled as multi-agent systems (MAS) in a centralized or distributed manner. In addition to the challenges in SARL such as long time horizon and sparse rewards, MARL brings extra challenges such as non-stationarity \cite{papoudakis2019dealing}, credit assignment \cite{nguyen2018credit} as well as scalability with the number of agents increasing \cite{hernandez2019survey, zhang2019multi}. Thus, platforms with flexible settings on the number of controlled agents become both important and valuable for relevant studies.

\subsubsection{Synchronized actions.}
Common SARL paradigm assumes that the environment will not change between when the environment state is observed and when the action is executed. The system is treated sequentially: the agent observes a state, freezes time while computing the action and applies the action, and then unfreezes time. The same settings normally apply to MARL, where multiple agents calculate actions together and then execute them at the same time step. However, it is usually not the case in the real world that the whole system's decision making and executing processes are synchronized. This results from agents' different reaction time and diverse action execution time under various situations, and it makes the whole MAS works asynchronously. For example, in multi-robot control areas, robots could behave asynchronously when executing different actions due to hardware limitations \cite{xiao2020thinking}.



\begin{figure*}[t]
\centering
\subfigure[]{
\begin{minipage}[t]{0.49\linewidth}
\centering
\includegraphics[width=1\columnwidth]{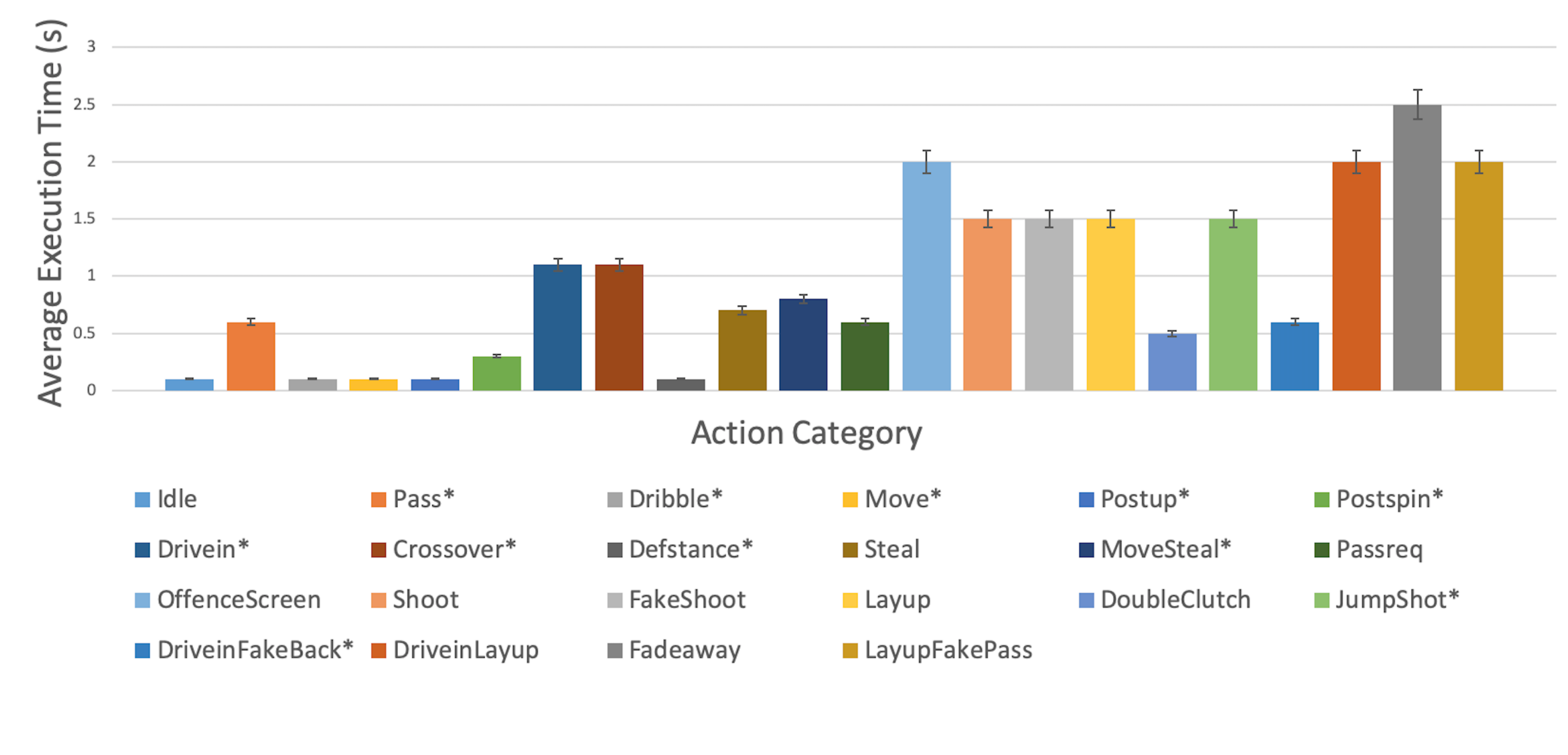}
\end{minipage}%
}%
\subfigure[]{
\begin{minipage}[t]{0.49\linewidth}
\centering
\includegraphics[width=1\columnwidth]{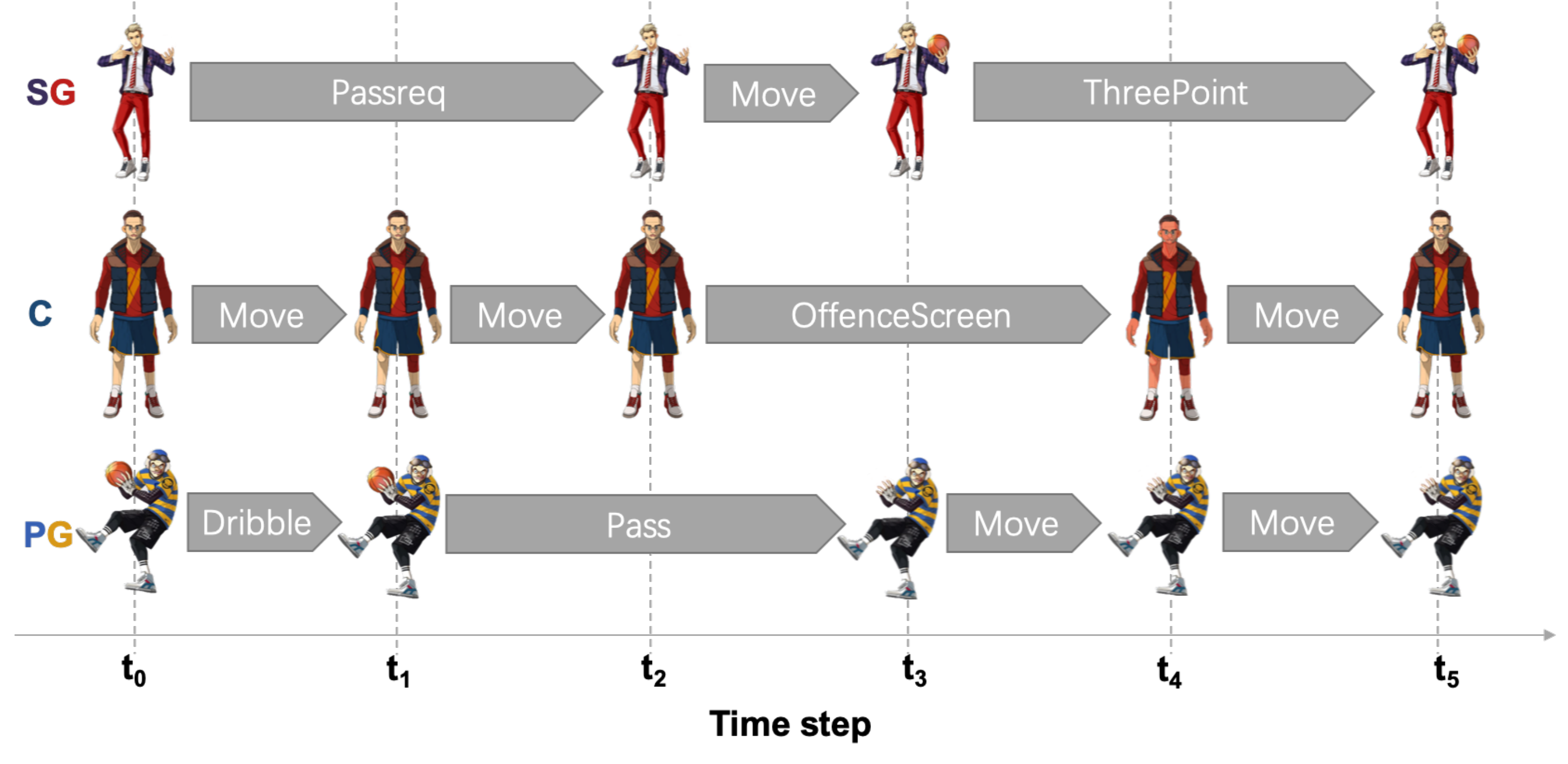}
\end{minipage}%
}%

\centering
\vspace{-10pt}
\setlength{\belowcaptionskip}{-10pt}
\caption{The asynchonized actions in Fever Basketball. (a) The average execution time of different action categories ($^*$ represents action units with different directions). (b) Illustration of the action asynchronism within the team.}
\label{async_action}
\end{figure*}

\subsubsection{Other related works.} 
There are also many other open-source environments focusing on certain game types and specific research fields. For example, the \textit{SMAC} \cite{vinyals2017starcraft}, which is a representative of the challenging RTS games, has been used as a test-bed for MARL algorithms despite that the additive and dense rewards settings could make it less challenging \cite{foerster2017counterfactual, rashid2018qmix, vinyals2019alphastar}. The \textit{DeepMind Control Suite} \cite{tassa2018deepmind}, \textit{AI2Thor} \cite{kolve2017ai2}, \textit{Habitat} \cite{savva2019habitat}, and \textit{PyBullet} \cite{coumans2016pybullet} environments are all related to continuous control tasks. The most similar platform to ours is the \textit{Google Research Football} \cite{kurach2019google}, which offers another kind of SPG: football. However, it is a synchronized game platform and there are many differences between football and basketball in terms of game settings like game rules, number of players as well as tactics.

\section{Fever Basketball Games}

Fever Basketball is an online basketball game, which simulates a half-court (length=11.4 meters, width=15 meters) basketball match between two teams\footnotemark \footnotetext[1]{https://github.com/FuxiRL/FeverBasketball}. The game includes the most common basketball aspects, such as jump ball, dribble, three-pointer, dunk, rebound, etc (see Figure 1\ref{basketball_rule} for a few examples). The objective of each team is to score as much as possible to win the match within a limited time.

\paragraph{Supported basketball elements.}

The game offers more than 30 characters (Charles, Alex, Steven, etc.) of different positions (C, PF, SF, PG, SG) with various attributes and skills to choose from before a match, which largely enriches both the randomness and challenges of the game.
At the beginning of each match, one player from each team will do the jump ball. The team which gets the ball will be the offense team and the other team will be the defense team. The player holding the ball in the offense team is in the state of \textit{attack} and the other two players are in the state of \textit{assist}. Players in the offense team can use offense actions (such as screen, fast break, jockey for position, etc.) and shooting actions (such as jump shot, layup, dunk, etc.) to score. Meanwhile, players in the defense team should try their best to prevent the offense team from scoring by applying defense strategy like one-on-one checks, steals, rejections, and so on. Once the ball is out of the hands of the possessed player such as after shoot or rebound, all of the players will be in the \textit{freeball} states. At such a moment, if players of the defense team manage to fetch the ball, they need to go through an \textit{attack-defense switch} process named \textit{ballclear} to prepare for offense by dribbling out of the three-point-line. Once the offense team scores, the ball possession will be handed to the opposite team. A typical match lasts for three minutes (with an average FPS of 60) except for the overtime. The shoot clock violation (20 seconds) will be punished by handling the ball possession to the opposite team.

\paragraph{Supported player control modes.}
Fever Basketball offers a convenient way to control game players by modifying corresponding keywords when launching the game clients. The number of players within each team can be chosen from $\{1, 2, 3\}$, which covers both the single-agent training and the multi-agent training scenarios with increasing complexity and difficulty in a curriculum manner. Meanwhile, the position of each player can be chosen from \{C, PF, SF, PG, SG\} and the game characters can be switched freely within the players we provide. Furthermore, the game also provides three control modes. The first one is the \textit{Bot} mode, where the agents can be trained with the built-in rule-based bots whose difficulty levels can be chosen from \{easy, medium, hard\} with different reaction time and shooting rate. The second one is the \textit{SelfPlay} mode, which allows the training of both teams through self-play. The third mode is the \textit{Human} mode, where the human player can control the specific position of the home team and fight against the built-in bots or pre-trained agents.



\paragraph{Game states and representations.}
Raw game states in Fever Basketball are data packages received from game clients, which includes information like current scene name (\textit{attack, assist, defense, freeball, ballclear}), general game information (such as attack remain time and scores), both teams' information (such as the player type, player position, facing angle and shoot rate), ball information (such as ball position, ball velocity, owned player and owned team) and the results of the last action. Please see the detailed description in the Appendix. Besides, we also provide a vector-based representation wrapper class corresponding to each game scene as well as some useful functions like distance calculation of two coordinates, based on which researchers can easily define their state representations for training.
We collect transition experience from 20 parallelled

\paragraph{Asynchronized game actions.}
To better simulate the real-world basketball game, one of the key features of Fever Basketball is that a player's primitive actions have different execution time (see Figure \ref{async_action}(a)), which makes the actions of the players within the same team asynchronized. For example, consider the offense scenario depicted in Figure \ref{async_action}(b), where the PG is dribbling the ball with the SG requesting the ball and the C keep moving at $t_0$. After getting the ball at time step $t_3$, which is passed from PG at time $t_1$, the SG makes a three-point shot that costs two time steps of execution (i.e., finishing shooting at $t_5$), with C keeps doing the OffenceScreen action (i.e. pick-and-roll) from $t_2$ to $t_4$.
Unlike common MARL environments which assume the agents' actions are synchronized, the asynchronism in Fever Basketball brings extra challenges for the current MARL algorithms, especially for the ones with centralized training \cite{sunehag2018value, rashid2018qmix}. Each agent faces a much more non-stationary environment since the other agents' ongoing actions may have a large impact on the state transitions observed by the agent.
Besides, the number of actions for different positions and different game scenes are listed in Table \ref{state&action_space} (3v3 mode). 
A detailed description of these actions can be found in Appendix.


\begin{table}[htbp]
\caption{Number of actions in Fever Basketball scenes.}
\centering
\resizebox{.9\columnwidth}{!}{
\begin{tabular}{c|c|c|c|c|c}
\hline
\hline
\diagbox{Scene}{Type} & C & PF & SF & PG & SG \\
\hline
Attack & 29 & 30 & 43 & 35 & 42 \\
Defense & 19 & 19 & 19 & 27 & 27 \\
Freeball & 10 & 10 & 9 & 9 & 9 \\
Ballclear & 22 & 22 & 25 & 25 & 25 \\
Assist & 11 & 11 & 11 & 11 & 11 \\
\hline
\end{tabular}
}
\label{state&action_space}
\end{table}
\vspace{-15pt}

\paragraph{Game rewards settings.}
Game rewards settings in Fever Basketball are also highly flexible and can be easily customized by researchers in terms of both the shaping rewards and the game rewards. We currently offer a set of game rewards related to corresponding game scenes. To be specific, in the \textit{offense} scene (\textit{attack \& assist}), the agent will be rewarded with 2 or 3 if the team goals while being punished with -1 if the ball is blocked, stolen, lost, or time is up. Rewards settings in the \textit{defense} scene are the opposite of the offense scene. For the \textit{freeball} scene, the agent will be rewarded with 1 if it possesses the ball and will be punished with -1 if it loses (such as that the opponent gets the ball or time is up). In the \textit{ballclear} scene, the agent will be rewarded with 1 if it gets the ball out of the three-point line successfully and will be punished similarly as those in the \textit{offense} scene.




\section{Fever Basketball Benchmarks}

Fever Basketball is a complex, flexible, and highly customizable game environment which allows researchers to try new ideas and solve problems in basketball games. Compared with the single-agent mode, the multi-agent mode is more challenging and includes both competition and collaboration scenarios. In addition, it brings new challenges such as asynchronized actions. To evaluate the performance of existing algorithms and handle the asynchronism problems in Fever Basketball, we propose some heuristic methods and provide a set of benchmarks regarding the 3v3 tasks. In all of these tasks, the goal of the trained agents is to score as many points as possible in a limited amount of time (3 minutes per round) against the built-in bots, whose difficulty levels range from easy, medium to hard. In addition, to facilitate fair comparisons, we use the SG position for both teams.

\begin{figure}[t]
\centering
\subfigure[Exp-Mask (Ms)]{
\begin{minipage}[t]{0.49\linewidth}
\centering
\includegraphics[width=0.98\columnwidth]{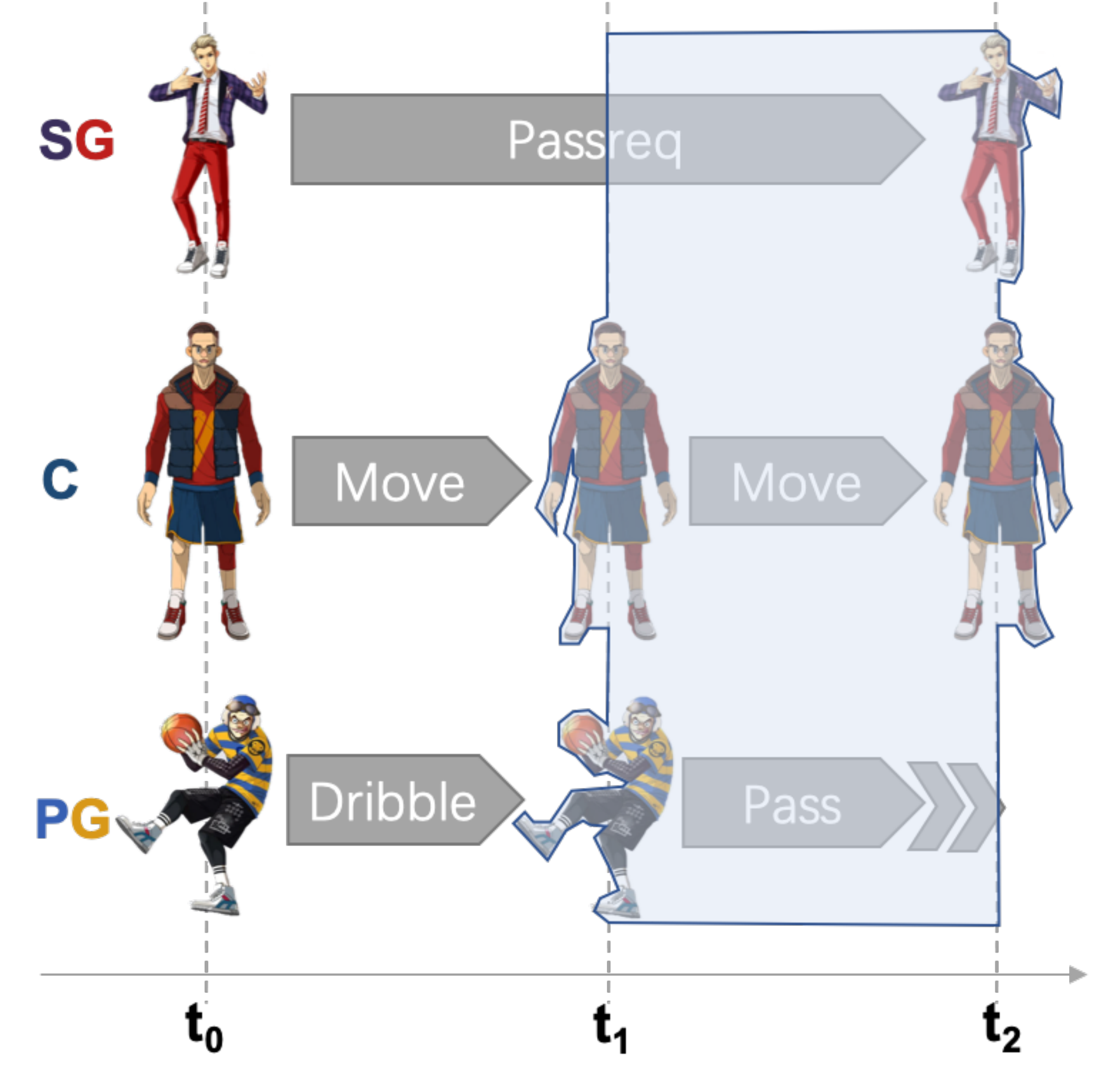}
\end{minipage}%
}%
\subfigure[Exp-Splice (Sp)]{
\begin{minipage}[t]{0.49\linewidth}
\centering
\includegraphics[width=0.98\columnwidth]{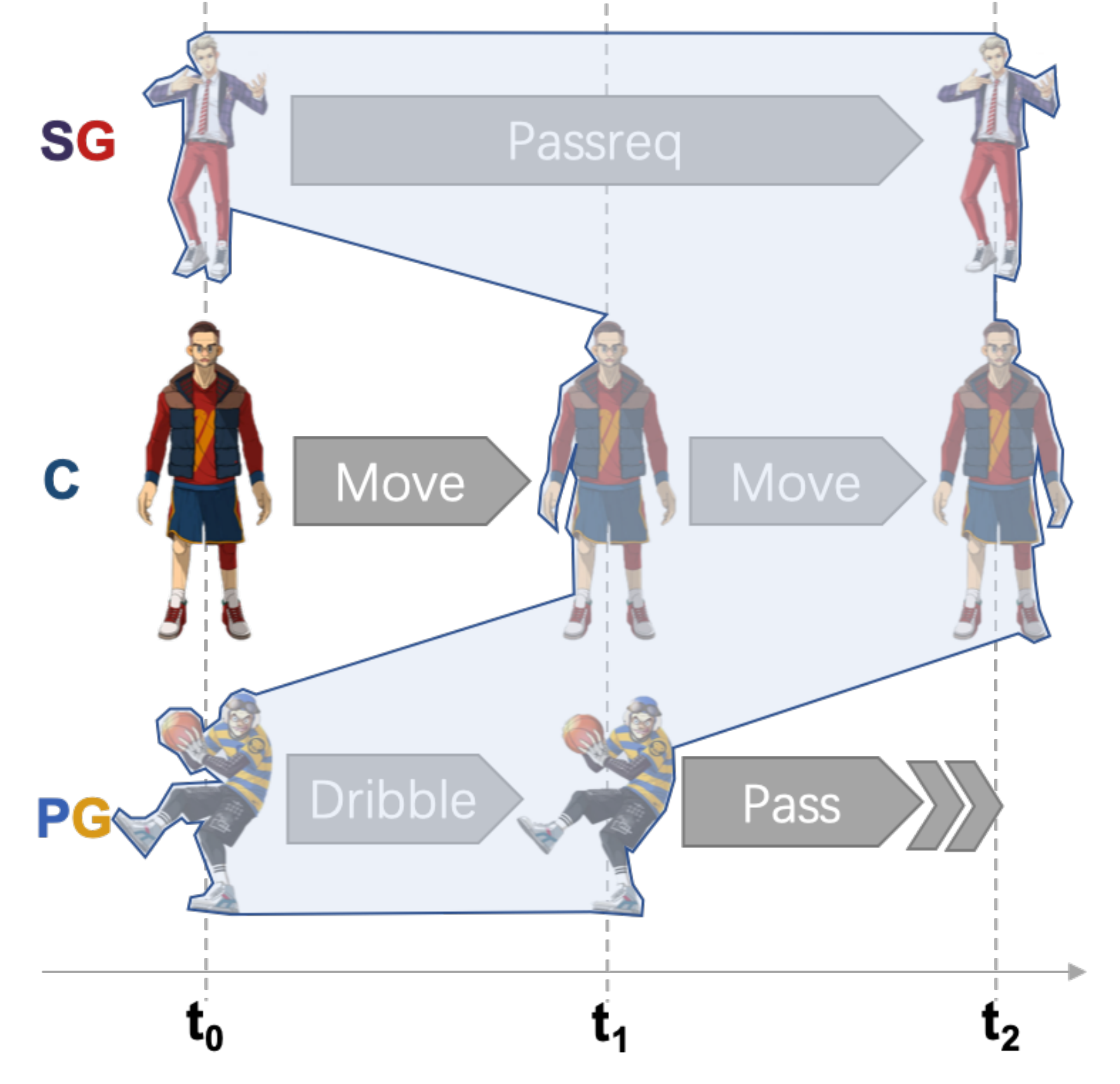}
\end{minipage}%
}%

\centering
\vspace{-5pt}
\setlength{\belowcaptionskip}{-10pt}
\caption{Proposed experience collection methods for JAL.}
\label{exp}
\end{figure}


\subsection{Methods}
Generally speaking, there are two major learning paradigms in MARL, namely the joint action learner (JAL) and the independent learner (IL) \cite{claus1998dynamics}. In cooperative settings, JAL, which also includes the \textit{centralised training with decentralised execution} paradigm, assumes all the agents' actions can be observed, such as VDN \cite{sunehag2018value} and QMIX \cite{rashid2018qmix}. In contrast, IL only relies on its action and the coordination can be achieved through heuristics of optimistic and average rewards, such as HYQ \cite{matignon2007hysteretic}, EXCEL\cite{hu2019explicitly}.

The modeling of asynchronized actions in Fever Basketball differs from that of \textit{MacDec-POMDPs} \cite{amato2019modeling, xiao2020macro} where \textit{options} are proposed, and applied to dynamic programming problems and model-free robot-control areas, respectively. Since the asynchronized actions in Fever Basketball are still primitive actions and the decision-making still focuses on a low level of granularity, and proper methods of collecting transitions remain critical.
In terms of experience collection, IL algorithms have advantages over JAL algorithms because their learning processes can be handled independently and do not rely on collecting other agents' on-going actions. However, it will be a problem to find an appropriate time to collect the joint-action transitions for JAL.


\begin{figure*}[t]
\centering
\begin{minipage}[t]{1\linewidth}
\centering
\includegraphics[width=0.98\columnwidth]{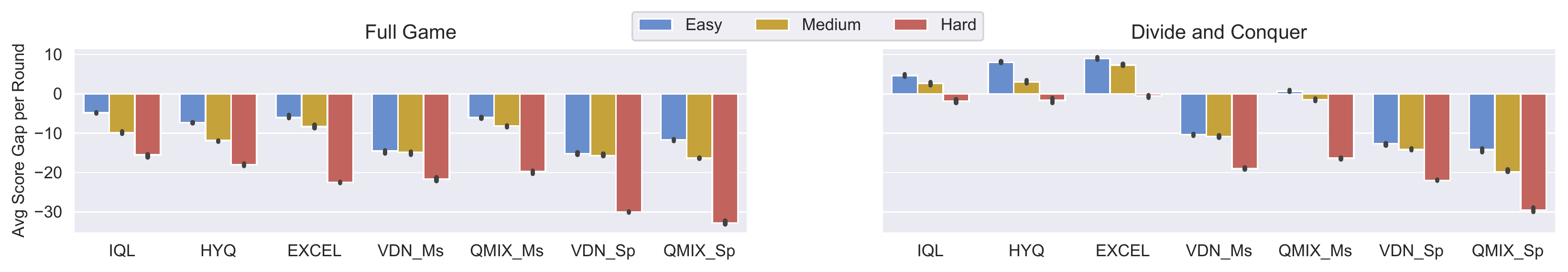}
\end{minipage}
\centering
\vspace{-10pt}
\setlength{\belowcaptionskip}{-10pt}
\caption{Benchmark experiments for Fever Basketball.}
\label{benchmark}
\end{figure*}

As illustrated in Figure \ref{exp}, we propose two methods to collect the joint-action experience. The first method is experience-mask (EXP-Ms), which masks the on-going actions out and regards them as \textit{Idle} when collecting joint transitions at a certain time-step. For example, if we denote $o_t^P$, $s_t^P$, $a_t^P$, $r_t$, $d_t$ as the global observation, local observation, action, global reward, and done information of player $P$ at time step $t$, respectively. The global experience in the shaded area of Figure \ref{async_action}(a) would be:
    \begin{align*}
      EXP_{Ms}=[o_{t_1}^C, (s_{t_1}^{SG}, s_{t_1}^C, s_{t_1}^{PG}), (Idle, a_{t_1}^C, a_{t_1}^{PG}), r_{t_2}, \\
        o_{t_2}^C, (s_{t_2}^{SG}, s_{t_2}^C, s_{t_2}^{PG}), {d}_{t_2}]
    \end{align*}
    
The second method is experience splice (EXP-Sp), which means that we collect the joint transition when all the players have finished the recent on-going actions, and then splice the experience to form final transitions based on the global states observed at the time step when the agents start to execute actions.
As illustrated by Figure \ref{async_action}(b), the joint transition experience in the shaded area would be:
\begin{align*}
    EXP_{Sp1}=[o_{t_0}^{SG}, (s_{t_0}^{SG}, s_{t_1}^C, s_{t_0}^{PG}), (a_{t_0}^{SG}, a_{t_1}^C, a_{t_1}^{PG}), r_{t_1\&t_2}, \\
    o_{t_2}^{SG}, (s_{t_2}^{SG}, s_{t_2}^C, s_{t_1}^{PG}), {d}_{t_1\&t_2}]
\\
    EXP_{Sp2}=[o_{t_1}^{\;\,C}, (s_{t_0}^{SG}, s_{t_1}^C, s_{t_0}^{PG}), (a_{t_0}^{SG}, a_{t_1}^C, a_{t_1}^{PG}), r_{t_1\&t_2}, \\
    o_{t_2}^{\;\,C}, (s_{t_2}^{SG}, s_{t_2}^C, s_{t_1}^{PG}), {d}_{t_1\&t_2}]
\\
    EXP_{Sp3}=[o_{t_1}^{\;\,C}, (s_{t_0}^{PG}, s_{t_1}^C, s_{t_0}^{PG}), (a_{t_0}^{SG}, a_{t_1}^C, a_{t_1}^{PG}), r_{t_1\&t_2}, \\
    o_{t_1}^{PG}, (s_{t_2}^{SG}, s_{t_2}^C, s_{t_1}^{PG}), {d}_{t_1\&t_2}]
\end{align*}
By learning from the reconstructed experience, we expect these two heuristic methods can help the joint-action learners acquire the perception of the execution time of corresponding actions to facilitate better coordination.

\subsection{Experimental results}

 In this section, we provide benchmark results for both IL algorithms IQL \cite{mnih2015human}, HYQ \cite{matignon2007hysteretic}, EXCEL \cite{hu2019explicitly}) and JAL algorithms VDN \cite{sunehag2018value}, QMIX \cite{rashid2018qmix} with the EXP-Ms and EXP-Sp methods. We evaluate these algorithms in both the \textit{Full Game} setting and \textit{Divide and Conquer} setting. In the \textit{Full Game} setting, the learners need to handle all the sub-tasks (i.e. \textit{offense (attack \& assist), defense, freeball, ballclear}) through one model with unavailable actions masked out. In the \textit{Divide and Conquer} setting, each of these sub-tasks is allocated with a corresponding learner, which decreases the difficulties of training. The technical details of the training architectures and hyperparameters can be found in Appendix.

The experimental results of the Fever Basketball Benchmarks, which are averaged over 10 game clients after trained for 100 rounds, are shown in Figure \ref{benchmark}. It can be found that the \textit{Full Game} setting is much more challenging than the \textit{Divide and Conquer} setting, where all of the algorithms fail to defeat the built-in bots. Besides, the hard bots are more difficult to be defeated than the medium and easy ones. The performances of the independent learners generally outperform the joint-action learners even though we try to eliminate the action asynchronism within the team by using EXP-Ms and EXP-Sp methods. In addition, it seems the EXP-Ms method performs relatively better than the EXP-Sp method, which might result from the neglect of some agent's short-time transitions when generating the global experience, such as the transition from $t_0$ to $t_1$ of player C in Figure \ref{exp}(b). The results indicate that the asynchronism problem is not well solved and worth further studying.

\begin{figure*}[t]
\centering
\includegraphics[width=2\columnwidth]{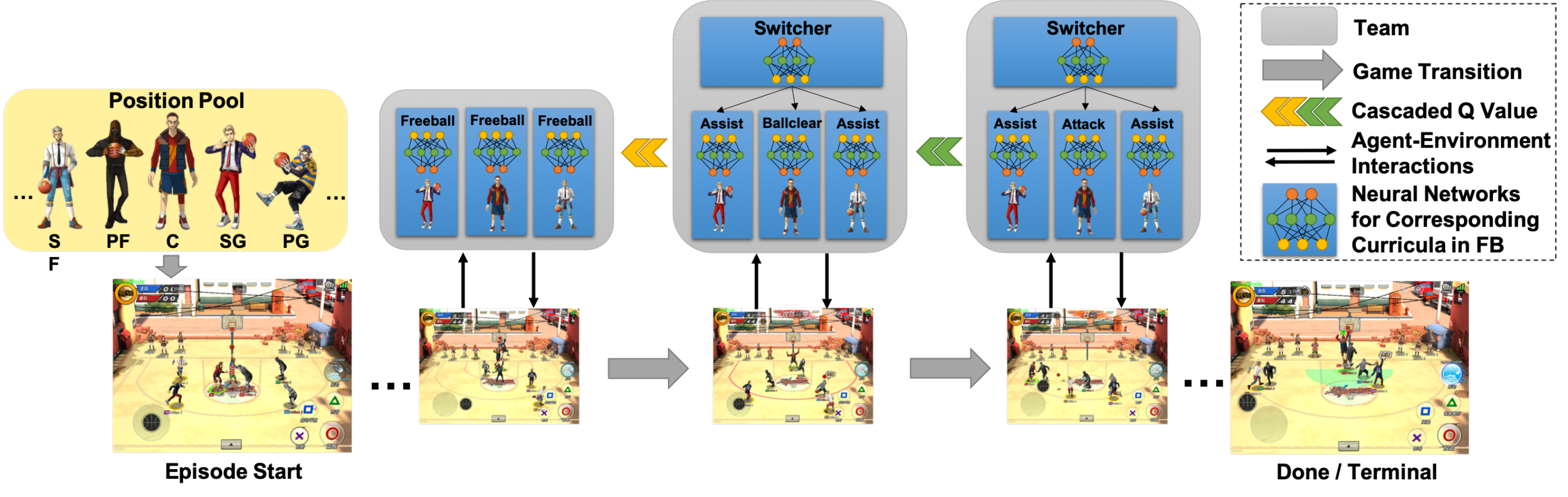}

\vspace{-10pt}
\setlength{\belowcaptionskip}{-10pt}
\caption{The Integrated Curricula Training framework (ICT).}
\label{basketball_models}
\end{figure*}

\section{The Integrated Curricula Training (ICT)}

Although the complex basketball problem can be partially solved through existing MARL algorithms under the \textit{Divide and Conquer} settings, the correlations between corresponding sub-tasks are neglected and there could be miscoordinations induced by the asynchronism. Besides, it also seems that the proposed Exp-Ms and Exp-Sp methods struggle to facilitate the learning of action execution time, and the asynchronism in Fever Basketball remains a critical problem, especially for joint-action learners. 
To make further progress, we take advantage of the independent learners and propose a curriculum learning based framework named ICT (Figure \ref{basketball_models}), which mainly includes a set of weighted cascading curricula learners and a coordination curricula switcher. These weighted cascading curricula learners are responsible for corresponding sub-tasks generated by basketball game rules. And the coordination curricula switcher, which has a relatively higher priority in making decisions, focuses on learning cooperative policy on primitive actions that will result in curriculum switch, such as the \textit{pass} action that triggers switch between the \textit{attack} and the \textit{assist} curricula.

\subsection{Methods}

\subsubsection{The weighted cascading curricula learners.}

Curriculum learning is used to solve complex and difficult problems \cite{wu2016training, wu2018master}. As mentioned before, Fever Basketball offers a set of base training scenarios according to game rules, namely \textit{attack}, \textit{defense}, \textit{freeball}, \textit{ballclear} and \textit{assist} from the perspective of a single player. All of these five base curricula are the fundamental aspects of an integrated basketball match. And only by mastering these basic curricula can one be ready for generating appropriate policies throughout the entire basketball match. Thus we intend to firstly train a corresponding DRL agent ($i$) to learn each of these base curricula ($\tau_i$), the interaction process of which can be formulated as a finite Markov Decision Process (MDP). During each episode, agent $i$ perceives the state of the corresponding base curriculum $s_t \in S_{\tau_i}$ at each time step \emph{t}, and outputs an action $a_t \in A_{\tau_i}$ according to policy $\pi_{\tau_i}$. A scalar reward $r_t \in R_{\tau_i}$ is then yielded from the environment and the agent will transit to a new state $s_{t+1}$ with a probability distribution $P(s_{t+1}|s_{t}, a_{t}, \tau_i)$. The transition $(s_t, a_t, r_t, s_{t+1})$ is stored in replay buffer $D_i$. Agent $i$ 's goal is to find an optimal policy $\pi^*$ to maximize the expected accumulative (discounted) rewards from each state s in corresponding curriculum $\tau_i$, namely the value function $Q^{*}_{\tau_i}(s,a)$ which can be formulated as:
    \begin{equation*}
        Q^{*}_{\tau_i}(s,a) = \mathbb{E}_{s^{'} \sim \varepsilon} [r + \gamma max_{a'} Q_{\tau_i}^*(s',a') | s, a, \tau_i]
    \end{equation*}
The update of the $Q_{\tau_i}$ network parameters $\theta_i$ are carried out by randomly sampling mini-batches from corresponding replay buffer $D_i$ and performing a gradient descent step on $(y_i - Q (s,a;\theta_i))^2$. The Q value labels $y_i$ can be calculated as follows:
	\begin{equation*}
	y_i= \begin{cases}
	r_i, & \text{terminal s of $\tau_i$} \\
	r_i + \gamma max_{a'} Q_{\tau_i}^{*}(s', a'; \theta_i), & \text{otherwise}
	\end{cases}
	\end{equation*}

In this way, the complicated and challenging basketball problem is decomposed into several easier curricula which can be preliminarily solved by applying co-training of corresponding DRL agents similar to the \textit{divide-and-conquer} strategy. Although these base curricula training has enabled the agents to acquire some primary skills towards corresponding sub-tasks, the whole basketball match remains a challenge. This is because a round of basketball match could include many inter-transitions between corresponding sub-tasks, and these five base curricula are actually highly correlated. For example, the \textit{attack} and \textit{assist} curricula are normally followed by \textit{freeball} after the shot of the offense team, thus the policy used in former curricula will contribute to the outcomes of the latter curricula.

To deal with the correlations between corresponding base curricula, we propose the cascading curricula training approach. It is implemented by adding the max $Q(s_{\tau_j}',a_{\tau_j}';\theta_{j})$ value of the following base curriculum $\tau_j$ to the reward $r_i$ that agent $i$ received from environment to form the new label $y_i$ when former base curriculum $\tau_i$ reaches a terminal. Meanwhile, we use a weight parameter $\eta \in [0,1]$ to adjust the ratio of the cascading Q values heuristically during training. The adjustment procedure for $\eta$ is crucial for both the stabilization and performance of the whole training process. When $\eta$ equals to 0, the cascading curricula training becomes the base curricula training. When $\eta$ increases to 1 by following certain heuristic procedures during training, the correlations between corresponding base curricula will be gradually established through the backup of the learned Q values and contribute to the final integrated policy throughout the whole basketball match. The new Q value labels $y_i^{cas}$ can be formulated as:
	\begin{equation*}
	y_i^{cas}= \begin{cases}
	r_i + \eta\gamma max_{a'} Q_{\tau_j}^{*}(s_{\tau_j}', a_{\tau_j}'; \theta_j), & \text{terminal s of $\tau_i$} \\
	r_i + \gamma max_{a'} Q_{\tau_i}^{*}(s', a'; \theta_i), & \text{otherwise}
	\end{cases}
	\end{equation*}

\subsubsection{The coordination curricula switcher.}

\begin{figure*}[t]
\centering
\subfigure[]{
\begin{minipage}[t]{0.33\linewidth}
\centering
\includegraphics[width=0.98\columnwidth]{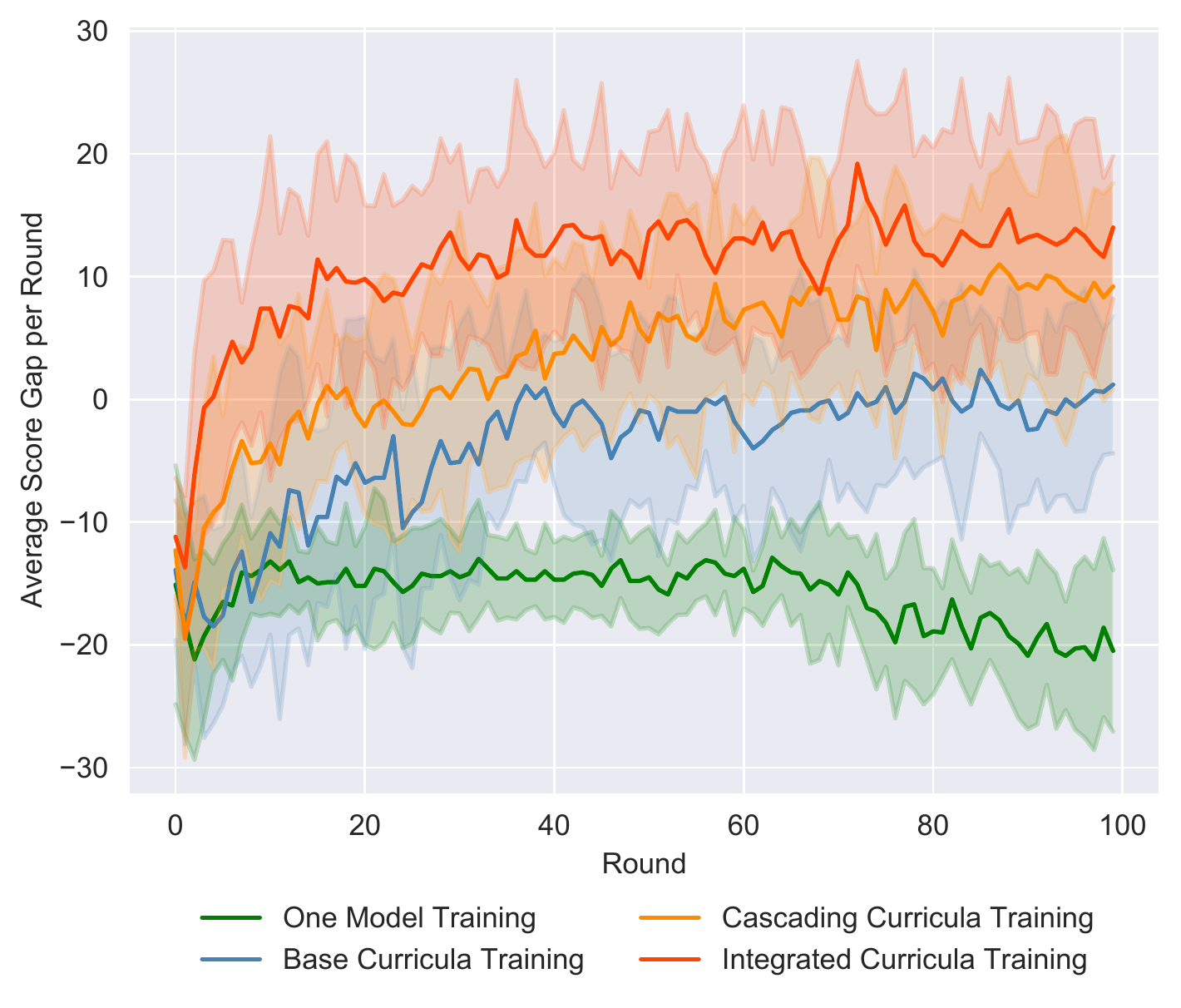}
\end{minipage}%
}%
\subfigure[]{
\begin{minipage}[t]{0.33\linewidth}
\centering
\includegraphics[width=0.98\columnwidth]{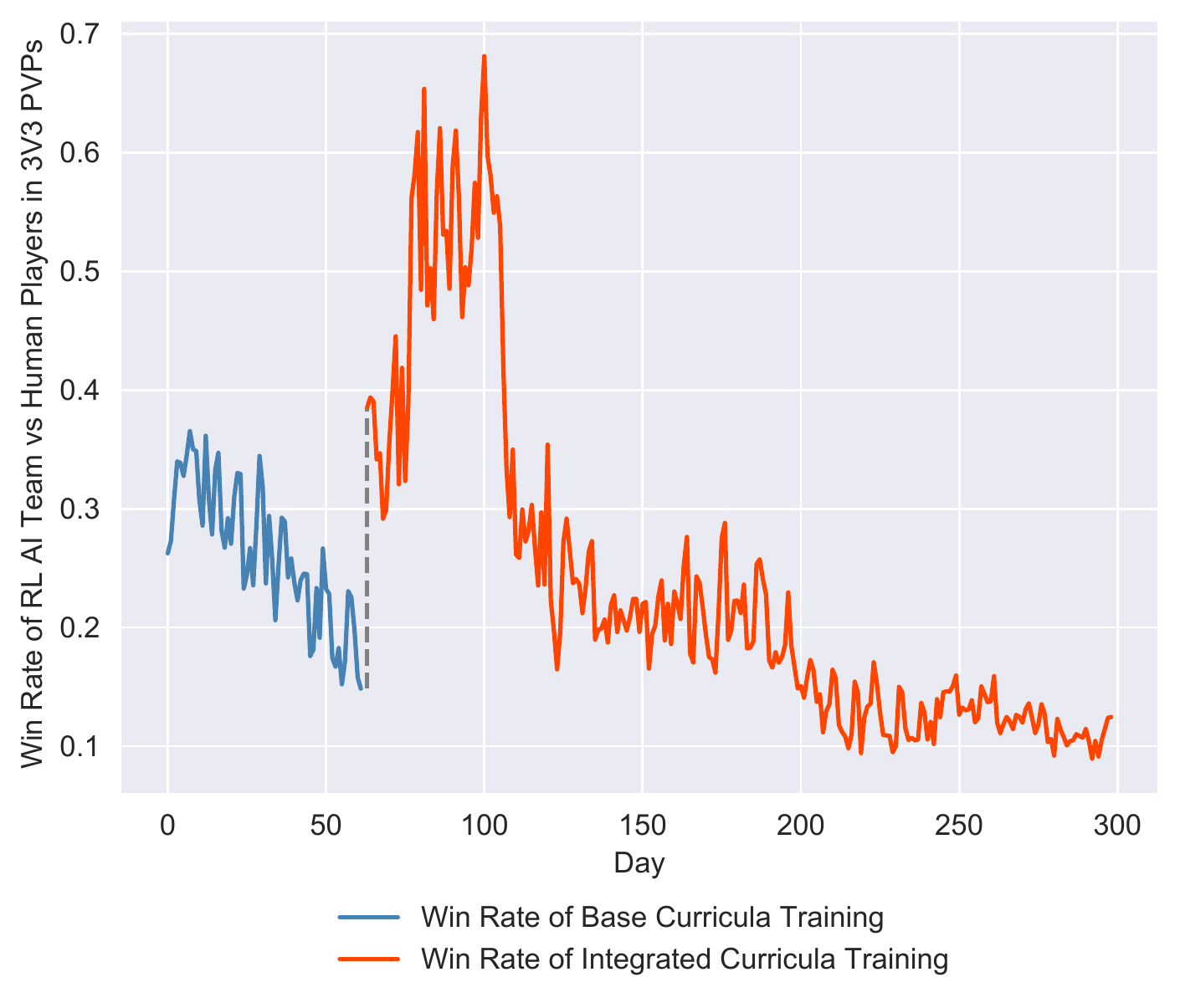}
\end{minipage}%
}%
\subfigure[]{
\begin{minipage}[t]{0.33\linewidth}
\centering
\includegraphics[width=0.98\columnwidth]{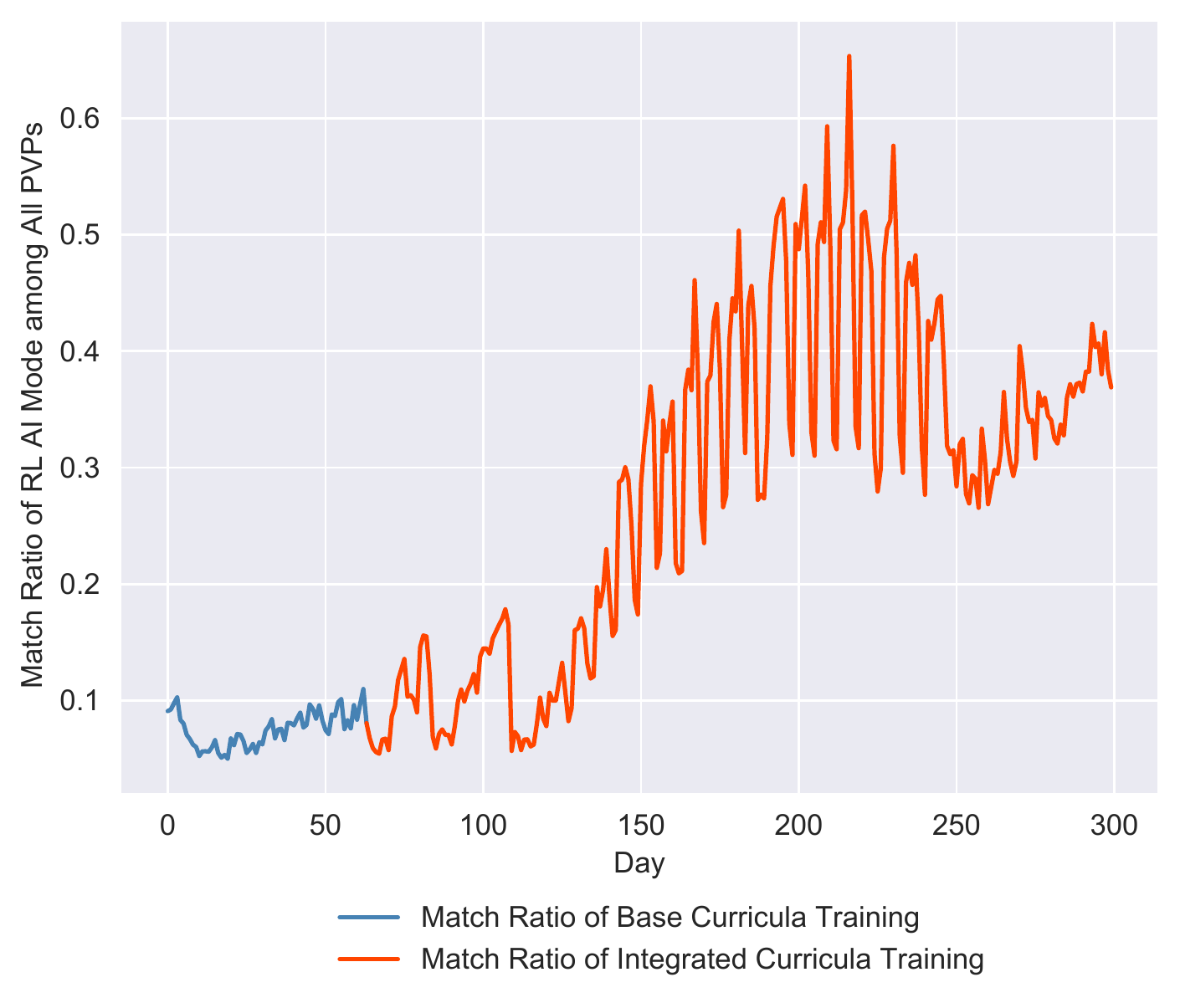}
\end{minipage}
}%
\centering
\vspace{-10pt}
\setlength{\belowcaptionskip}{-10pt}
\caption{Performance of training approaches proposed in Fever Basketball. (a) Evaluation with hard built-in bots. (b) Win-rate of self-play models during online evaluation with human players. (c) Match-ratio of self-play models during online evaluation with human players.}
\label{basketball_results}
\end{figure*}

Although the weighted cascading curricula training can degrade the complex basketball problem into relatively easier base curricula and take their correlations into account, it is only from the perspective of a single player. However, as a typical team sport, coordination within the same team plays a crucial role in all basketball matches. Based on the cascading curricula training, we propose a high-level coordination curricula switcher to facilitate the training of coordination by focusing on learning cooperative actions that could induce curricula switching within the same team. For example, the \textit{pass} action, which is the core primary action that transfers ball possession and creates basketball tactics in the offense team. By taking over such action, the coordination curricula switcher will focus on learning how to pass the ball to the right player in an appropriate time, which, in the meantime, will also result in the curriculum switch between \textit{attack} / \textit{ballclear} and \textit{assist} within the same team. The coordination curricula switcher will have a relatively higher priority over those weighted cascading base curricula learners on action selection to ensure the performance of coordination when it is necessary. Meanwhile, by taking over coordination related actions from original action spaces, it also enables a reduction of original action spaces for \textit{attack}, \textit{ballclear} and \textit{assist} scene, which helps the agents to learn policies more effectively in corresponding curricula. The pseudo-code of the ICT framework for Fever Basketball can be found in the Appendix (see Algorithm $1$).


\subsection{Experimental results}
	
	


In this subsection, an ablation study is firstly carried out to assess the effects of different parts in our ICT framework by playing with the hard built-in bots, which includes the one-model training, the base curricula training, the cascading curricula training, and ICT framework. The performance of the whole ICT framework is then further evaluated with online players in Fever Basketball. The APEX-Rainbow \cite{horgan2018distributed, hessel2018rainbow} algorithm is used for all learners and we put the detailed training setups and architectures in Appendix.

The results of the ablation experiments are demonstrated in Figure \ref{basketball_results}(a). The horizontal axis is the number of evaluated matches along with the training process (3 minutes per round). The vertical axis is the average score gap between proposed training approaches and the built-in hard bots in one match over 10 game clients. We can find that the one-model training approach performs the worst and the players trained by this approach struggle to master these five distinct sub-tasks together. The base curriculum training approach performs much better than the one-model training method since it can focus only on the corresponding sub-task and generate some fundamental policies towards basic game rules. Players trained in this way tend to play solo while lacking tactical movements since correlations between related sub-tasks are ignored. The weighted cascading curriculum training can make further improvements compared with the base curriculum training because the correlation between related sub-tasks is retained and the policy can be optimized over the whole task despite that the coordination within one team remains a weakness. The ICT framework significantly outperforms other training approaches since the coordination performance can be essentially improved by using the coordination curricula switcher.


The results of a 300-day online evaluation with $1,130,926$ human players are illustrated in Figure \ref{basketball_results}(b and c). During this evaluation, we first test the model learned by the base curriculum learning and change the online model to the one trained through the ICT framework on day 63. As is shown in Figure \ref{basketball_results}(b), the win rate of the updated model (almost up to 70\%, red broken lines) increases more than twice of the former model (around 30\%, blue broken lines) at the beginning of each online evaluations with human players in 3v3 PVP (i.e. player vs player) matches. The team trained with our method can generate many professional coordination tactics like give-and-go, and it is more likely to create wide-open areas to score by passing smoothly. In addition, the match ratio that human players participated in playing against the challenging AI teams keeps increasing (see Figure \ref{basketball_results} (c)) among all the PVP matches, which indicates that we bring extra revenues for the game.

\section{Discussion and Conclusion}
In this paper, we present the Fever Basketball Environment, a novel open-source reinforcement learning environment of the basketball game. It is a complex and challenging environment which supports both single-agent training and multi-agent training. Besides, the actions with different execution time in this environment make it a good platform for studying the challenging problem of asynchronized multi-agent decision-making. We implement and evaluate the state-of-the-art MARL algorithms (such as VDN, Qmix, and EXCEL) together with two heuristic methods (i.e. EXP-Ms and EXP-Sp) to alleviate the effect of asynchronism in both the \textit{Full Game} setting and \textit{Divide and Conquer} setting of Fever Basketball. The results show that the game is challenging and existing algorithms fail to solve the asynchronism problems. To shed light on this complex task, we take advantage of the curriculum learning and propose an integrated curricula training framework to solve this problem step by step. Though progress has been made, the win-rate against on-line human players is not high (up to 70\%) and keeps decreasing as the evaluation process goes, which may result from our model's lacking generation to unseen opponents, and meanwhile demonstrates the difficulties of mastering the basketball game. We expect the components involved in Fever Basketball such as the complexity, the flexible settings , and the asynchronism will be useful for investigating current scientific challenges like long-time horizon, spare rewards, credit assignment, non-stationarity.

\section{Ethical Impact}
Considering that the game platforms have substantially boosted the development of reinforcement learning (RL), the open-source of our Fever Basketball platform is expected to further enrich the types of existing virtual environments for RL communities. What's more, the new challenges brought by our platform are also of great potential to incubate new algorithms, which is another aspect for contributing to the development of RL.


\bigskip

\bibliography{Fever_Basketball.bib}

\begin{thebibliography}{47}
\providecommand{\natexlab}[1]{#1}
\providecommand{\url}[1]{\texttt{#1}}
\providecommand{\urlprefix}{URL }
\expandafter\ifx\csname urlstyle\endcsname\relax
  \providecommand{\doi}[1]{doi:\discretionary{}{}{}#1}\else
  \providecommand{\doi}{doi:\discretionary{}{}{}\begingroup
  \urlstyle{rm}\Url}\fi

\bibitem[{Amato et~al.(2019)Amato, Konidaris, Kaelbling, and
  How}]{amato2019modeling}
Amato, C.; Konidaris, G.; Kaelbling, L.~P.; and How, J.~P. 2019.
\newblock Modeling and planning with macro-actions in decentralized POMDPs.
\newblock \emph{Journal of Artificial Intelligence Research} 64: 817--859.

\bibitem[{Badia et~al.(2020)Badia, Piot, Kapturowski, Sprechmann, Vitvitskyi,
  Guo, and Blundell}]{badia2020agent57}
Badia, A.~P.; Piot, B.; Kapturowski, S.; Sprechmann, P.; Vitvitskyi, A.; Guo,
  D.; and Blundell, C. 2020.
\newblock Agent57: Outperforming the atari human benchmark.
\newblock \emph{arXiv preprint arXiv:2003.13350} .

\bibitem[{Beattie et~al.(2016)Beattie, Leibo, Teplyashin, Ward, Wainwright,
  K{\"u}ttler, Lefrancq, Green, Vald{\'e}s, Sadik et~al.}]{beattie2016deepmind}
Beattie, C.; Leibo, J.~Z.; Teplyashin, D.; Ward, T.; Wainwright, M.;
  K{\"u}ttler, H.; Lefrancq, A.; Green, S.; Vald{\'e}s, V.; Sadik, A.; et~al.
  2016.
\newblock Deepmind lab.
\newblock \emph{arXiv preprint arXiv:1612.03801} .

\bibitem[{Bellemare et~al.(2013)Bellemare, Naddaf, Veness, and
  Bowling}]{bellemare2013arcade}
Bellemare, M.~G.; Naddaf, Y.; Veness, J.; and Bowling, M. 2013.
\newblock The arcade learning environment: An evaluation platform for general
  agents.
\newblock \emph{Journal of Artificial Intelligence Research} 47: 253--279.

\bibitem[{Claus and Boutilier(1998)}]{claus1998dynamics}
Claus, C.; and Boutilier, C. 1998.
\newblock The dynamics of reinforcement learning in cooperative multiagent
  systems.
\newblock \emph{AAAI/IAAI} 1998(746-752): 2.

\bibitem[{Cobbe et~al.(2019)Cobbe, Hesse, Hilton, and
  Schulman}]{cobbe2019leveraging}
Cobbe, K.; Hesse, C.; Hilton, J.; and Schulman, J. 2019.
\newblock Leveraging procedural generation to benchmark reinforcement learning.
\newblock \emph{arXiv preprint arXiv:1912.01588} .

\bibitem[{Coumans and Bai(2016)}]{coumans2016pybullet}
Coumans, E.; and Bai, Y. 2016.
\newblock Pybullet, a python module for physics simulation for games, robotics
  and machine learning .

\bibitem[{Foerster et~al.(2017)Foerster, Farquhar, Afouras, Nardelli, and
  Whiteson}]{foerster2017counterfactual}
Foerster, J.; Farquhar, G.; Afouras, T.; Nardelli, N.; and Whiteson, S. 2017.
\newblock Counterfactual multi-agent policy gradients.
\newblock \emph{arXiv preprint arXiv:1705.08926} .

\bibitem[{Haarnoja et~al.(2018)Haarnoja, Zhou, Abbeel, and
  Levine}]{haarnoja2018soft}
Haarnoja, T.; Zhou, A.; Abbeel, P.; and Levine, S. 2018.
\newblock Soft actor-critic: Off-policy maximum entropy deep reinforcement
  learning with a stochastic actor.
\newblock \emph{arXiv preprint arXiv:1801.01290} .

\bibitem[{Heinrich and Silver(2016)}]{heinrich2016deep}
Heinrich, J.; and Silver, D. 2016.
\newblock Deep reinforcement learning from self-play in imperfect-information
  games.
\newblock \emph{arXiv preprint arXiv:1603.01121} .

\bibitem[{Hernandez-Leal, Kartal, and Taylor(2019)}]{hernandez2019survey}
Hernandez-Leal, P.; Kartal, B.; and Taylor, M.~E. 2019.
\newblock A survey and critique of multiagent deep reinforcement learning.
\newblock \emph{Autonomous Agents and Multi-Agent Systems} 33(6): 750--797.

\bibitem[{Hessel et~al.(2018)Hessel, Modayil, Van~Hasselt, Schaul, Ostrovski,
  Dabney, Horgan, Piot, Azar, and Silver}]{hessel2018rainbow}
Hessel, M.; Modayil, J.; Van~Hasselt, H.; Schaul, T.; Ostrovski, G.; Dabney,
  W.; Horgan, D.; Piot, B.; Azar, M.; and Silver, D. 2018.
\newblock Rainbow: Combining improvements in deep reinforcement learning.
\newblock In \emph{Thirty-Second AAAI Conference on Artificial Intelligence}.

\bibitem[{Horgan et~al.(2018)Horgan, Quan, Budden, Barth-Maron, Hessel,
  Van~Hasselt, and Silver}]{horgan2018distributed}
Horgan, D.; Quan, J.; Budden, D.; Barth-Maron, G.; Hessel, M.; Van~Hasselt, H.;
  and Silver, D. 2018.
\newblock Distributed prioritized experience replay.
\newblock \emph{arXiv preprint arXiv:1803.00933} .

\bibitem[{Hu et~al.(2019)Hu, Chen, Fan, and Hao}]{hu2019explicitly}
Hu, Y.; Chen, Y.; Fan, C.; and Hao, J. 2019.
\newblock Explicitly Coordinated Policy Iteration.
\newblock In \emph{IJCAI}, 357--363.

\bibitem[{Jiang, Ekwedike, and Liu(2018)}]{jiang2018feedback}
Jiang, D.~R.; Ekwedike, E.; and Liu, H. 2018.
\newblock Feedback-based tree search for reinforcement learning.
\newblock \emph{arXiv preprint arXiv:1805.05935} .

\bibitem[{Juliani et~al.(2018)Juliani, Berges, Vckay, Gao, Henry, Mattar, and
  Lange}]{juliani2018unity}
Juliani, A.; Berges, V.-P.; Vckay, E.; Gao, Y.; Henry, H.; Mattar, M.; and
  Lange, D. 2018.
\newblock Unity: A general platform for intelligent agents.
\newblock \emph{arXiv preprint arXiv:1809.02627} .

\bibitem[{Juliani et~al.(2019)Juliani, Khalifa, Berges, Harper, Teng, Henry,
  Crespi, Togelius, and Lange}]{juliani2019obstacle}
Juliani, A.; Khalifa, A.; Berges, V.-P.; Harper, J.; Teng, E.; Henry, H.;
  Crespi, A.; Togelius, J.; and Lange, D. 2019.
\newblock Obstacle tower: A generalization challenge in vision, control, and
  planning.
\newblock \emph{arXiv preprint arXiv:1902.01378} .

\bibitem[{Kolve et~al.(2017)Kolve, Mottaghi, Han, VanderBilt, Weihs, Herrasti,
  Gordon, Zhu, Gupta, and Farhadi}]{kolve2017ai2}
Kolve, E.; Mottaghi, R.; Han, W.; VanderBilt, E.; Weihs, L.; Herrasti, A.;
  Gordon, D.; Zhu, Y.; Gupta, A.; and Farhadi, A. 2017.
\newblock Ai2-thor: An interactive 3d environment for visual ai.
\newblock \emph{arXiv preprint arXiv:1712.05474} .

\bibitem[{Kurach et~al.(2019)Kurach, Raichuk, Sta{\'n}czyk, Zaj{\k{a}}c,
  Bachem, Espeholt, Riquelme, Vincent, Michalski, Bousquet
  et~al.}]{kurach2019google}
Kurach, K.; Raichuk, A.; Sta{\'n}czyk, P.; Zaj{\k{a}}c, M.; Bachem, O.;
  Espeholt, L.; Riquelme, C.; Vincent, D.; Michalski, M.; Bousquet, O.; et~al.
  2019.
\newblock Google research football: A novel reinforcement learning environment.
\newblock \emph{arXiv preprint arXiv:1907.11180} .

\bibitem[{Lample and Chaplot(2017)}]{lample2017playing}
Lample, G.; and Chaplot, D.~S. 2017.
\newblock Playing FPS games with deep reinforcement learning.
\newblock In \emph{Thirty-First AAAI Conference on Artificial Intelligence}.

\bibitem[{Liu et~al.(2019)Liu, Zheng, Li, Bian, and Song}]{ijcai2019-631}
Liu, T.; Zheng, Z.; Li, H.; Bian, K.; and Song, L. 2019.
\newblock Playing Card-Based RTS Games with Deep Reinforcement Learning.
\newblock In \emph{Proceedings of the Twenty-Eighth International Joint
  Conference on Artificial Intelligence, {IJCAI-19}}, 4540--4546. International
  Joint Conferences on Artificial Intelligence Organization.
\newblock \doi{10.24963/ijcai.2019/631}.
\newblock \urlprefix\url{https://doi.org/10.24963/ijcai.2019/631}.

\bibitem[{Matignon, Laurent, and Le~Fort-Piat(2007)}]{matignon2007hysteretic}
Matignon, L.; Laurent, G.~J.; and Le~Fort-Piat, N. 2007.
\newblock Hysteretic q-learning: an algorithm for decentralized reinforcement
  learning in cooperative multi-agent teams.
\newblock In \emph{2007 IEEE/RSJ International Conference on Intelligent Robots
  and Systems}, 64--69. IEEE.

\bibitem[{Mnih et~al.(2013)Mnih, Kavukcuoglu, Silver, Graves, Antonoglou,
  Wierstra, and Riedmiller}]{mnih2013playing}
Mnih, V.; Kavukcuoglu, K.; Silver, D.; Graves, A.; Antonoglou, I.; Wierstra,
  D.; and Riedmiller, M. 2013.
\newblock Playing atari with deep reinforcement learning.
\newblock \emph{arXiv preprint arXiv:1312.5602} .

\bibitem[{Mnih et~al.(2015)Mnih, Kavukcuoglu, Silver, Rusu, Veness, Bellemare,
  Graves, Riedmiller, Fidjeland, Ostrovski et~al.}]{mnih2015human}
Mnih, V.; Kavukcuoglu, K.; Silver, D.; Rusu, A.~A.; Veness, J.; Bellemare,
  M.~G.; Graves, A.; Riedmiller, M.; Fidjeland, A.~K.; Ostrovski, G.; et~al.
  2015.
\newblock Human-level control through deep reinforcement learning.
\newblock \emph{Nature} 518(7540): 529.

\bibitem[{Munemasa et~al.(2018)Munemasa, Tomomatsu, Hayashi, and
  Takagi}]{munemasa2018deep}
Munemasa, I.; Tomomatsu, Y.; Hayashi, K.; and Takagi, T. 2018.
\newblock Deep reinforcement learning for recommender systems.
\newblock In \emph{2018 International Conference on Information and
  Communications Technology (ICOIACT)}, 226--233. IEEE.

\bibitem[{Nguyen, Kumar, and Lau(2018)}]{nguyen2018credit}
Nguyen, D.~T.; Kumar, A.; and Lau, H.~C. 2018.
\newblock Credit assignment for collective multiagent RL with global rewards.
\newblock In \emph{Advances in Neural Information Processing Systems},
  8102--8113.

\bibitem[{Nichol et~al.(2018)Nichol, Pfau, Hesse, Klimov, and
  Schulman}]{nichol2018gotta}
Nichol, A.; Pfau, V.; Hesse, C.; Klimov, O.; and Schulman, J. 2018.
\newblock Gotta learn fast: A new benchmark for generalization in rl.
\newblock \emph{arXiv preprint arXiv:1804.03720} .

\bibitem[{OpenAI(2018)}]{OpenAI_dota}
OpenAI. 2018.
\newblock OpenAI Five.
\newblock \url{https://blog.openai.com/openai-five/}.

\bibitem[{Paine et~al.(2019)Paine, Gulcehre, Shahriari, Denil, Hoffman, Soyer,
  Tanburn, Kapturowski, Rabinowitz, Williams et~al.}]{paine2019making}
Paine, T.~L.; Gulcehre, C.; Shahriari, B.; Denil, M.; Hoffman, M.; Soyer, H.;
  Tanburn, R.; Kapturowski, S.; Rabinowitz, N.; Williams, D.; et~al. 2019.
\newblock Making Efficient Use of Demonstrations to Solve Hard Exploration
  Problems.
\newblock \emph{arXiv preprint arXiv:1909.01387} .

\bibitem[{Pan et~al.(2017)Pan, You, Wang, and Lu}]{pan2017virtual}
Pan, X.; You, Y.; Wang, Z.; and Lu, C. 2017.
\newblock Virtual to real reinforcement learning for autonomous driving.
\newblock \emph{arXiv preprint arXiv:1704.03952} .

\bibitem[{Papoudakis et~al.(2019)Papoudakis, Christianos, Rahman, and
  Albrecht}]{papoudakis2019dealing}
Papoudakis, G.; Christianos, F.; Rahman, A.; and Albrecht, S.~V. 2019.
\newblock Dealing with non-stationarity in multi-agent deep reinforcement
  learning.
\newblock \emph{arXiv preprint arXiv:1906.04737} .

\bibitem[{Rashid et~al.(2018)Rashid, Samvelyan, De~Witt, Farquhar, Foerster,
  and Whiteson}]{rashid2018qmix}
Rashid, T.; Samvelyan, M.; De~Witt, C.~S.; Farquhar, G.; Foerster, J.; and
  Whiteson, S. 2018.
\newblock QMIX: Monotonic value function factorisation for deep multi-agent
  reinforcement learning.
\newblock \emph{arXiv preprint arXiv:1803.11485} .

\bibitem[{Savva et~al.(2019)Savva, Kadian, Maksymets, Zhao, Wijmans, Jain,
  Straub, Liu, Koltun, Malik et~al.}]{savva2019habitat}
Savva, M.; Kadian, A.; Maksymets, O.; Zhao, Y.; Wijmans, E.; Jain, B.; Straub,
  J.; Liu, J.; Koltun, V.; Malik, J.; et~al. 2019.
\newblock Habitat: A platform for embodied ai research.
\newblock In \emph{Proceedings of the IEEE International Conference on Computer
  Vision}, 9339--9347.

\bibitem[{Schulman et~al.(2017)Schulman, Wolski, Dhariwal, Radford, and
  Klimov}]{schulman2017proximal}
Schulman, J.; Wolski, F.; Dhariwal, P.; Radford, A.; and Klimov, O. 2017.
\newblock Proximal policy optimization algorithms.
\newblock \emph{arXiv preprint arXiv:1707.06347} .

\bibitem[{Shalev-Shwartz, Shammah, and Shashua(2016)}]{shalev2016safe}
Shalev-Shwartz, S.; Shammah, S.; and Shashua, A. 2016.
\newblock Safe, multi-agent, reinforcement learning for autonomous driving.
\newblock \emph{arXiv preprint arXiv:1610.03295} .

\bibitem[{Silver et~al.(2016)Silver, Huang, Maddison, Guez, Sifre, Van
  Den~Driessche, Schrittwieser, Antonoglou, Panneershelvam, Lanctot
  et~al.}]{silver2016mastering}
Silver, D.; Huang, A.; Maddison, C.~J.; Guez, A.; Sifre, L.; Van Den~Driessche,
  G.; Schrittwieser, J.; Antonoglou, I.; Panneershelvam, V.; Lanctot, M.;
  et~al. 2016.
\newblock Mastering the game of Go with deep neural networks and tree search.
\newblock \emph{nature} 529(7587): 484.

\bibitem[{Silver et~al.(2017)Silver, Schrittwieser, Simonyan, Antonoglou,
  Huang, Guez, Hubert, Baker, Lai, Bolton et~al.}]{silver2017mastering}
Silver, D.; Schrittwieser, J.; Simonyan, K.; Antonoglou, I.; Huang, A.; Guez,
  A.; Hubert, T.; Baker, L.; Lai, M.; Bolton, A.; et~al. 2017.
\newblock Mastering the game of go without human knowledge.
\newblock \emph{Nature} 550(7676): 354.

\bibitem[{Sunehag et~al.(2018)Sunehag, Lever, Gruslys, Czarnecki, Zambaldi,
  Jaderberg, Lanctot, Sonnerat, Leibo, Tuyls et~al.}]{sunehag2018value}
Sunehag, P.; Lever, G.; Gruslys, A.; Czarnecki, W.~M.; Zambaldi, V.~F.;
  Jaderberg, M.; Lanctot, M.; Sonnerat, N.; Leibo, J.~Z.; Tuyls, K.; et~al.
  2018.
\newblock Value-Decomposition Networks For Cooperative Multi-Agent Learning
  Based On Team Reward.
\newblock In \emph{AAMAS}, 2085--2087.

\bibitem[{Sutton and Barto(2018)}]{sutton2018reinforcement}
Sutton, R.~S.; and Barto, A.~G. 2018.
\newblock \emph{Reinforcement learning: An introduction}.
\newblock MIT press.

\bibitem[{Tassa et~al.(2018)Tassa, Doron, Muldal, Erez, Li, Casas, Budden,
  Abdolmaleki, Merel, Lefrancq et~al.}]{tassa2018deepmind}
Tassa, Y.; Doron, Y.; Muldal, A.; Erez, T.; Li, Y.; Casas, D. d.~L.; Budden,
  D.; Abdolmaleki, A.; Merel, J.; Lefrancq, A.; et~al. 2018.
\newblock Deepmind control suite.
\newblock \emph{arXiv preprint arXiv:1801.00690} .

\bibitem[{Vinyals et~al.(2019)Vinyals, Babuschkin, Chung, Mathieu, Jaderberg,
  Czarnecki, Dudzik, Huang, Georgiev, Powell et~al.}]{vinyals2019alphastar}
Vinyals, O.; Babuschkin, I.; Chung, J.; Mathieu, M.; Jaderberg, M.; Czarnecki,
  W.~M.; Dudzik, A.; Huang, A.; Georgiev, P.; Powell, R.; et~al. 2019.
\newblock AlphaStar: Mastering the real-time strategy game StarCraft II.
\newblock \emph{DeepMind Blog} .

\bibitem[{Vinyals et~al.(2017)Vinyals, Ewalds, Bartunov, Georgiev, Vezhnevets,
  Yeo, Makhzani, K{\"u}ttler, Agapiou, Schrittwieser
  et~al.}]{vinyals2017starcraft}
Vinyals, O.; Ewalds, T.; Bartunov, S.; Georgiev, P.; Vezhnevets, A.~S.; Yeo,
  M.; Makhzani, A.; K{\"u}ttler, H.; Agapiou, J.; Schrittwieser, J.; et~al.
  2017.
\newblock Starcraft ii: A new challenge for reinforcement learning.
\newblock \emph{arXiv preprint arXiv:1708.04782} .

\bibitem[{Wu and Tian(2016)}]{wu2016training}
Wu, Y.; and Tian, Y. 2016.
\newblock Training agent for first-person shooter game with actor-critic
  curriculum learning .

\bibitem[{Wu, Zhang, and Song(2018)}]{wu2018master}
Wu, Y.; Zhang, W.; and Song, K. 2018.
\newblock Master-Slave Curriculum Design for Reinforcement Learning.
\newblock In \emph{IJCAI}, 1523--1529.

\bibitem[{Xiao et~al.(2020)Xiao, Jang, Kalashnikov, Levine, Ibarz, Hausman, and
  Herzog}]{xiao2020thinking}
Xiao, T.; Jang, E.; Kalashnikov, D.; Levine, S.; Ibarz, J.; Hausman, K.; and
  Herzog, A. 2020.
\newblock Thinking While Moving: Deep Reinforcement Learning with Concurrent
  Control.
\newblock \emph{arXiv preprint arXiv:2004.06089} .

\bibitem[{Xiao, Hoffman, and Amato(2020)}]{xiao2020macro}
Xiao, Y.; Hoffman, J.; and Amato, C. 2020.
\newblock Macro-Action-Based Deep Multi-Agent Reinforcement Learning.
\newblock \emph{arXiv preprint arXiv:2004.08646} .

\bibitem[{Zhang, Yang, and Ba{\c{s}}ar(2019)}]{zhang2019multi}
Zhang, K.; Yang, Z.; and Ba{\c{s}}ar, T. 2019.
\newblock Multi-agent reinforcement learning: A selective overview of theories
  and algorithms.
\newblock \emph{arXiv preprint arXiv:1911.10635} .

\end{thebibliography}

\end{document}